\pdfoutput=1

\documentclass[11pt,dvipsnames]{article}

\usepackage{acl}
\usepackage{xcolor}
\usepackage{times}
\usepackage{latexsym}
\usepackage{rotating}
\usepackage{multirow,graphicx}
\usepackage{verbatim}
\usepackage[T1]{fontenc}

\usepackage[utf8]{inputenc}

\usepackage{microtype}

\newcommand{\dataset}{\texttt{kp-biomed}}
\newcommand{\kptimes}{\texttt{KPTimes}}
\newcommand{\kpk}{\texttt{KP20k}}
\newcommand{\namedkeys}{\texttt{NamedKeys}}

\usepackage{booktabs}
\usepackage{pgfplots}
\pgfplotsset{compat=newest}

\title{A Large-Scale Dataset for Biomedical Keyphrase Generation}

\author{Maël Houbre, Florian Boudin \and Béatrice Daille \\
         Nantes Université, École Centrale Nantes, CNRS, LS2N, UMR 6004, F-44000 Nantes, France\\
         \texttt{first.last@univ-nantes.fr}}

\begin{document}
\maketitle
\begin{abstract}
Keyphrase generation is the task consisting in generating a set of words or phrases that highlight the main topics of a document. There are few datasets for keyphrase generation in the biomedical domain and they do not meet the expectations in terms of size for training generative models. In this paper, we introduce \dataset{}, the first large-scale biomedical keyphrase generation dataset with more than 5M documents collected from PubMed abstracts. We train and release several generative models and conduct a series of experiments showing that using large scale datasets improves significantly the performances for present and absent keyphrase generation. The dataset is available under CC-BY-NC v4.0 license at \url{https://huggingface.co/datasets/taln-ls2n/kpbiomed}.
\end{abstract}

\section{Introduction}

Keyphrase generation aims at automatically generating a set of keyphrases, that is, words and phrases that summarize a given document.
Since they distill the important information from documents, keyphrases have showed to be useful in many applications, most notably in information retrieval~\cite{10.1145/42005.42016,zhai-1997-fast,10.1145/312624.312671,10.1145/1141753.1141800,boudin-etal-2020-keyphrase} and summarization~\cite{10.1145/564376.564398,wan-etal-2007-towards,qazvinian-etal-2010-citation}.

Current models for generating keyphrases are built upon the sequence-to-sequence architecture~\cite{NIPS2014_a14ac55a} and are able to generate absent keyphrases that is, keyphrases that do not appear in the source text. However, training these models require large amounts of labeled data~\cite{meng-etal-2021-empirical}.
Unfortunately, such data is only available for limited domains and languages which greatly limits the applicability of these models~\cite{ye-wang-2018-semi}.
This work addresses this issue and introduces \dataset{}, the first large-scale dataset for keyphrase generation in the biomedical domain.

Creating labeled data for keyphrase generation is a challenging task, requiring expert annotators and great effort~\cite{kim-etal-2010-semeval,augenstein-etal-2017-semeval}.
A commonly-used approach to cope with this task is to collect scientific abstracts and use keyphrases provided by authors as a proxy for expert annotations.
Authors provide keyphrases without any vocabulary constraint to highlight important points of their article; whereas indexers use a specific vocabulary and focus on indexing the article within a collection \citep{neveol2010author}. Therefore, keyphrases may differ from MeSH headings which are another indexing resource in the biomedical domain.
Fortunately, author keyphrases are becoming increasingly available in the biomedical domain~\cite{neveol2010author}, since they can be incorporated into search strategies in \href{https://www.nlm.nih.gov/pubs/techbull/jf13/jf13_pm_keywords.html}{PubMed} to improve retrieval effectiveness~\cite{https://doi.org/10.1002/asi.22985}.
Despite this, the largest keyphrase-labeled biomedical dataset that we know of has about 3k abstracts, all of which are labeled with present-only keyphrases~\cite{10.1145/3307339.3342147}.
In this paper, we take advantage of the expansive PubMed database to build a sufficiently large dataset to train biomedical keyphrase generation models\footnote{\kpk{} is currently considered as the reference dataset size (\(\geq 500k\)) to train keyphrase generation models}. We then compare models trained with different training set sizes to highlight the impact of dataset sizes in keyphrase generation.
Our contributions are as follows:
\begin{itemize}
    \item \dataset{}, a large, publicly available dataset for keyphrase generation in the biomedical domain, available through the Huggingface dataset platform\footnote{\href{https://huggingface.co/datasets/taln-ls2n/kpbiomed}{https://huggingface.co/datasets/taln-ls2n/kpbiomed}}; 
    \item Transformer-based models for biomedical keyphrase generation, providing open benchmarks to stimulate further work in the area\footnote{\href{https://huggingface.co/datasets/taln-ls2n/kpbiomed}{https://huggingface.co/datasets/taln-ls2n/kpbiomed-models}};
    \item Performance analysis of our models, which provides valuable insights into their generalization ability to other domains.
\end{itemize}

\section{Dataset}

\begin{table*}[htbp!]
\centering
\pgfplotsset{
    petitHist/.style={
        ticks=none,
        axis line style={draw=none},
        tick style={draw=none},
        ymin=0, ymax=100,
        xmin=0.5, xmax=2.5,
        width=2.4cm, height=1.9cm,
        ybar=3pt,
        bar shift=0pt,
    }
}

\begin{tabular}{clrrrrrcc}
\toprule
    \textbf{Domain} & \textbf{Dataset} &
    \textbf{\#train} & \textbf{\#val} & \textbf{\#test} & 
    \textbf{\#doc len} &  \textbf{\#kp} &  \textbf{\#kp len}  & 
    \begin{tikzpicture}
        \begin{axis}[petitHist]
            \addplot [draw=OrangeRed, fill=OrangeRed!50, text=OrangeRed] coordinates {(1, 100)};
            \addplot [draw=Green, fill=Green!70, text=SeaGreen] coordinates {(2, 100)};
            \node[text=black] at (1,50) {P};
            \node[text=black] at (2,50) {A};
        \end{axis}
    \end{tikzpicture}
    \\
\midrule
\vspace{1mm}

     & \textbf{\dataset{}} (ours)  & 
    5.6M & 20k & 20k & 
    271 & 5.3 & 1.9  &
    \begin{tikzpicture}
        \begin{axis}[petitHist]
            \addplot [draw=OrangeRed, fill=OrangeRed!50, text=OrangeRed] coordinates {(1, 65.8)}; 
            \addplot [draw=Green, fill=Green!70, text=SeaGreen] coordinates {(2, 34.2)};
            
        \end{axis}
    \end{tikzpicture} \\

    Biomedical & NamedKeys &
    -- & -- & 3k & 
    276 &  14.3 & 1.9  &
    \begin{tikzpicture}
        \begin{axis}[petitHist]
            \addplot [draw=OrangeRed, fill=OrangeRed!50, text=OrangeRed] coordinates {(1, 90.9)}; 
            \addplot [draw=Green, fill=Green!70, text=SeaGreen] coordinates {(2, 9.1)}; 

            
        \end{axis}
    \end{tikzpicture}
    \\

    & Schutz & 
    -- & -- & 1.3k &
    5.4k &  5.4 &  1.9 &
    \begin{tikzpicture}
        \begin{axis}[petitHist]
            \addplot [draw=OrangeRed, fill=OrangeRed!50, text=OrangeRed] coordinates {(1, 84.5)};
            \addplot [draw=Green, fill=Green!70, text=SeaGreen] coordinates {(2, 15.5)};

        \end{axis}
    \end{tikzpicture} \\

\midrule
    \multirow{2}{*}{General}  & 
    KP20k & 
    530k & 20k & 20k &
    175 &  5.3 & 2.1 &
    \begin{tikzpicture}
        \begin{axis}[petitHist]
            \addplot [draw=OrangeRed, fill=OrangeRed!50, text=OrangeRed] coordinates {(1, 58.2)};
            \addplot [draw=Green, fill=Green!70, text=SeaGreen] coordinates {(2, 41.8)};

        \end{axis}
    \end{tikzpicture} \\
    
    \multirow{2}{*}{scientific}  & SemEval-2010 & 
    144 & -- & 100 &
    192 &  15.4 & 2.1  &
    \begin{tikzpicture}
        \begin{axis}[petitHist]
            \addplot [draw=OrangeRed, fill=OrangeRed!50, text=OrangeRed] coordinates {(1, 42.16)};
            \addplot [draw=Green, fill=Green!70, text=SeaGreen] coordinates {(2, 57.84)};

        \end{axis}
    \end{tikzpicture} \\
    
   \multirow{2}{*}{articles}  & Inspec & 
    1k & 500 & 500 &
    138 &  9.8 &  2.3 &
    \begin{tikzpicture}
        \begin{axis}[petitHist]
            \addplot [draw=OrangeRed, fill=OrangeRed!50, text=OrangeRed] coordinates {(1, 78.00)};
            \addplot [draw=Green, fill=Green!70, text=SeaGreen] coordinates {(2, 22.00)};

        \end{axis}
    \end{tikzpicture} \\
    

    & LDKP10k &
    1.3M & 10k & 10k &
    4.9k & 6.9 & 2.1 &
    \begin{tikzpicture}
        \begin{axis}[petitHist]
            \addplot [draw=OrangeRed, fill=OrangeRed!50, text=OrangeRed] coordinates {(1, 63.65)};
            \addplot [draw=Green, fill=Green!70, text=SeaGreen] coordinates {(2, 36.35)};
            
        \end{axis}
    \end{tikzpicture} \\

\midrule

    News&
     KPTimes & 
    260k & 10k & 20k &
    921 & 5.0  & 1.5  &
    \begin{tikzpicture}
        \begin{axis}[petitHist]
            \addplot [draw=OrangeRed, fill=OrangeRed!50, text=OrangeRed] coordinates {(1, 45.61)};
            \addplot [draw=Green, fill=Green!70, text=SeaGreen] coordinates {(2, 54.39)};

        \end{axis}
    \end{tikzpicture} \\

\bottomrule
\end{tabular}








\caption{Statistics of the proposed dataset. For comparison purposes, we also report statistics of commonly-used and other biomedical datasets. Columns \texttt{P} and \texttt{A} are respectively the percentage of keyphrases occurring in the source text and absent ones.}
\label{tab:statistics}
\end{table*}

We employ the December 2021 baseline set of MEDLINE/PubMed citation records\footnote{\href{https://ftp.ncbi.nlm.nih.gov/pubmed/baseline/}{https://ftp.ncbi.nlm.nih.gov/pubmed/baseline/}} as a resource for collecting abstracts, which contains over 33 million records. We extracted all the records (5.9 million) that include a title, an abstract and some author keyphrases. Records of papers published between 1939 and 2011 only account for a small fraction of these extracted records (3\%) and were further filtered out to avoid possible diachronic issues. Last, we went through the remaining records to split the semicolon-separated list of author keyphrases and discard those having keyphrases with punctuation in it. The resulting dataset is composed of 5.6 million abstracts and was randomly and evenly divided by publishing year into training, validation and test splits.
To investigate the impact of the amount of training data on the quality of the generated keyphrases, the training split was further divided into increasingly large subsets: small (500k), medium (2M) and large (5.6M). The training splits are also evenly divided by publishing year.

Statistics of the \dataset{} dataset are detailed in Table~\ref{tab:statistics} along with other commonly-used datasets for keyphrase generation and extraction.
%
%
We are aware of only two datasets in the biomedical domain: \namedkeys{}~\citep{10.1145/3307339.3342147} which is made up of MEDLINE/PubMed abstracts and is therefore mostly included in \dataset{}, and Schutz~\citep{schutz2008keyphrase} which is composed of full-text articles from the same source. It is worth noting that these datasets are very limited in size (3k and 1.3k documents respectively) compared to recent keyphrase generation datasets \kpk{}~\citep{meng-etal-2017-deep}, \kptimes{}~\citep{gallina-etal-2019-kptimes} and \texttt{LDKP10k}~\citep{mahata_ldkp_2022}. 
Table~\ref{tab:statistics} shows that thanks to the amount of papers available in MEDLINE/PubMed, \dataset{} is the largest of all aforementioned datasets, being more than 10 times larget than \kpk{} which is the current reference dataset for keyphrase generation. The average number of keyphrases per document (\#kp) in \dataset{} is roughly the same than in \kpk{} and \texttt{LDKP10k} which have their keyphrases assigned by authors as well. However, we see that this number is way below the average number of keyphrases assigned by professional indexers like in Inspec~\citep{hulth-2003-improved} or when authors' keyphrases are combined with readers' as in SemEval-2010~\citep{kim-etal-2010-semeval}. The unusually high number of keyphrases per document in \namedkeys{}, despite having author assigned keyphrases, is because of two restrictive criteria. Indeed, each article has at least 5 keyphrases all of which have to occur in the source text. The average number of words per keyphrase (\#kp\_len) is also comparable for all scientific datasets regardless of the kind of annotators.

Using keyphrases as proxies for indexing or expanding documents with queries composed of words that do not appear in the source text, has been proven more useful to enhance document retrieval than using words occurring in the text \citep{boudin-etal-2020-keyphrase,nogueira_document_2019}. In keyphrase generation, we call those keyphrases absent keyphrases, for which several definitions are being used. We refer to the definition from \cite{meng-etal-2017-deep} “we denote phrases that do not match any contiguous subsequence of source text as absent keyphrases" which was then precised in \cite{boudin-gallina-2021-redefining}. In \cite{10.1145/3307339.3342147} the keyphrase "anesthesia" is considered present if the word "postanesthesia" is in the source text. In our case, it is considered absent which is why \namedkeys{} does not appear with 100\% present keyphrases in Table~\ref{tab:statistics}.
The main difference between \dataset{} and \namedkeys{}, despite the number of documents, is the proportion of absent keyphrases. \dataset{} contains about 34\% of absent keyphrases which is in the same range as scientific datasets \kpk{} and \texttt{LDKP10k} that were designed to train neural generative approaches~\citep{meng-etal-2017-deep,mahata_ldkp_2022}.
\section{Experiments}

\subsection{Models}
In keyphrase generation, the architectures are currently mainly based on autoencoders with Recurrent Neural Networks~\citep{meng-etal-2017-deep,chen_keyphrase_2018,chen_title-guided_2019,chan_neural_2019} or Transformers~\citep{meng-etal-2021-empirical,ahmad-etal-2021-select}. 

Following the work of \cite{meng-etal-2021-empirical} that obtained state-of-the-art results with Transformers, we used two different generative BART models~\citep{lewis_bart_2020} and compared their performances on different domains. However, in this article we did not seek to get state-of-the-art results, but rather introduce \dataset{} to the community with results on well known baselines, which is why we employed pre-trained models that we just fine-tuned for keyphrase generation~\citep{chowdhury_applying_2022}. The models are BioBART-base~\citep{yuan-etal-2022-biobart} which is already pre-trained on PubMed and BART-base~\citep{lewis_bart_2020} which is pre-trained on news, books and webtext. To the best of our knowledge, there is no generic scientific BART model. Therefore, we chose BioBART for fine-tuning on scientific datasets rather than BART. Models are available via the huggingface platform.

For comparison with extractive approaches, we considered MultipartiteRank \cite{boudin_unsupervised_2018} as a baseline, which is state-of-the-art in unsupervised graph-based keyphrase extraction. We used the implementation available in the keyphrase extraction toolkit \texttt{pke}\footnote{\href{https://github.com/boudinfl/pke}{https://github.com/boudinfl/pke}} with the default settings.

\subsection{Experimental settings}
We followed the One2Seq paradigm~\cite{meng-etal-2021-empirical} for training which consists of generating the keyphrases of an input article as a single sequence. For each article, we concatenated the ground truth keyphrases as a single sequence with a special delimiter. Following \cite{meng-etal-2021-empirical}, present keyphrases were ordered by their first occurrence in the source text followed by the absent ones.

We trained each model for 10 epochs with a batch size of 128. We set the input length limit at 512 tokens for the text and 128 tokens for the reference keyphrase sequence. All the parameters and the training were handled with the huggingface trainer API\footnote{Our code is available for reproducibility. \href{https://github.com/MHoubre/src-kpbiomed}{https://github.com/MHoubre/kpbiomed} }. Hyperparameters and hardware details are available in appendix~\ref{sec:training_settings}. Training the BioBART-base model on the small training split for 10 epochs took about 9 hours and about 110 hours on the large training split. Once models were trained, we over-generated keyphrase sequences using beam search with a beam width of 20 for evaluation. Inference on test sets took around 50 minutes each.

\begin{table*}[htbp!]
\centering
\begin{tabular}{lcccccccc}
\hline
 Model  & \multicolumn{2}{ c }{\dataset} & \multicolumn{2}{ c }{KP20k} & \multicolumn{2}{ c }{KPTimes} \\

\cmidrule(lr){2-3} \cmidrule(lr){4-5} \cmidrule(lr){6-7} 

& F1@10 & F1@M & F1@10 & F1@M & F1@10 & F1@M \\
\hline

MultipartiteRank   & 15.3 & -- & 12.9 & -- & 16.7 & --  \\

BioBART-small  & 31.4 & 32.5 & 25.2 & 27.1 &  22.0 & 24.4 \\

BioBART-medium   & \underline{32.5}$\textsuperscript{\dag}$ & \underline{33.8}$\textsuperscript{\dag}$ & 26.2$\textsuperscript{\dag}$ & 28.2$\textsuperscript{\dag}$ & 22.1  & 24.6  \\

BioBART-large  & \textbf{33.1}$\textsuperscript{\dag}$ & \textbf{34.7}$\textsuperscript{\dag}$ & \underline{26.9}$\textsuperscript{\dag}$ & \underline{28.9}$\textsuperscript{\dag}$ & \underline{23.5}$\textsuperscript{\dag}$  & \underline{26.2}$\textsuperscript{\dag}$  \\

BioBART-KP20k & 28.2 & 29.5 & \textbf{28.6}$\textsuperscript{\dag}$ & \textbf{31.9}$\textsuperscript{\dag}$ & 16.8 & 19.2 \\

BART-KPTimes & 9.1 & 9.6 & 3.6 & 2.7 & \textbf{29.7}$\textsuperscript{\dag}$ & \textbf{39.4}$\textsuperscript{\dag}$ \\

\hline

\end{tabular}
\caption{\label{present_models-results}
Performances of the models on present keyphrase generation. \dag means significant improvements over BioBART-small. Second best results are underlined.
}
\end{table*}

\subsection{Evaluation}

We evaluated our models on 3 datasets, \dataset{} for biomedical data, \kpk{} for generic scientific documents and \kptimes{} for news articles. We did not use \namedkeys{} as a test set as we noticed a substantial overlap with our training set. We evaluated present and absent keyphrase generation separately to get better insights of our models' performances. To that end, we only compared each model's output to the present (respectively absent) keyphrases of the ground truth. For present keyphrases we employed F1@M and F1@10. F1@M is the F1 measure applied on the first keyphrase sequence generated by the model whereas F1@10 evaluates the top ten generated keyphrases. We evaluated absent keyphrase generation with R@10 which is the recall on the top 10 generated keyphrases. As F1@10 and R@10 require 10 keyphrases, if we did not have enough unique keyphrases with our over generation, we added the token "<unk>" until we reached 10 keyphrases. The generated keyphrases and the reference were stemmed with the Porter Stemmer to reduce matching errors. 
To measure statistical significance, we opted for Student's t-test at p < 0.01. 

\subsection{Results}

The macro-averaged results of the evaluation are reported in Table~\ref{present_models-results} and Table~\ref{absent_models-results}. BioBART-KP20k (respectively BART-KPTimes) stands for the BioBART (respectively BART) model which has been fine-tuned on \kpk{} (respectively \kptimes{}). For BioBART models, we add the size of the \dataset{} training split in the name for clarity.

\begin{table}[htbp]
\centering
\resizebox{\columnwidth}{!}{
\begin{tabular}{lcccc}
\hline
 Model  & \dataset{} & KP20k & KPTimes \\

\cmidrule(lr){2-4} 

& R@10 & R@10 & R@10 \\
\hline

BioBART-small  & 3.3 & 1.8 & 2.6 \\

BioBART-medium   & \underline{3.6}$\textsuperscript{\dag}$ & \underline{1.9} & \underline{2.7}  \\

BioBART-large  & \textbf{4.1}$\textsuperscript{\dag}$ & \underline{1.9} & 2.1 \\

BioBART-KP20k & 2.9 & \textbf{5.5}$\textsuperscript{\dag}$ & 1.6 \\

BART-KPTimes & 1.5 & 0.8 & \textbf{39.1}$\textsuperscript{\dag}$ \\

\hline

\end{tabular}
}
\caption{\label{absent_models-results}
Performances of the models on absent keyphrase generation. \dag means significant improvements over BioBART-small. Second best results are underlined.
}
\end{table}
Transformer based approaches achieve the best results but only on the datasets they were trained on as previously showed for RNN based approaches in \cite{gallina-etal-2019-kptimes}. 
For present keyphrase generation, BioBART-large achieves significant improvements compared to its small and medium counterparts in all datasets. This shows that using more data does improve the performances of the generative approaches in predicting present keyphrases in in and out of domain data. The performance drop of BioBART-KP20K on \dataset{} is interestingly much more controlled than BioBART models' on KP20k. Compared to BioBART-small which has been trained on the same amount of data, the drop in F1@M is only of 7.5\% relative for BioBART-KP20k when it is of 16.6\% relative for BioBART-small. We think that BioBART's pretraining may be beneficial for BioBART-KP20k on \dataset{}. On news articles though, BioBART-KP20k shows a relative drop of 35\%, when it is only of 25\% relative for BioBART-small. When used on out of domain data, BART-KPTimes performs even worse than MultipartiteRank.  

In absent keyphrase generation, models fail in attaining significant improvements outside of their domain. Using more data does not seem to help for out of domain absent keyphrase generation. We can explain the high results of BART-KPTimes on its test set by the fact that many of the absent keyphrases are common to numerous articles. 

We also think that the keyphrase order that we chose for training is one reason for the models' poor abstractive results. To verify this hypothesis, we compute the average percentage of the models' predictions appearing in the source text. Results are reported in Table~\ref{ext-results}. For @10, we removed all the added <unk> tokens before computing. It is clear that the extraction percentage of each model decreases when using top 10 predictions on all datasets. This shows that models prioritize generating present keyphrases which can then lead to low quality absent candidates.

\begin{table}[htbp!]
\centering
\resizebox{\columnwidth}{!}{
\begin{tabular}{lcccccc}
\hline
 Model & \multicolumn{2}{ c }{\dataset} & \multicolumn{2}{ c }{KP20k} & \multicolumn{2}{ c }{KPTimes}  \\

\cmidrule(lr){2-3} \cmidrule(lr){4-5} \cmidrule(lr){6-7}  

  & @M & @10 & @M & @10 & @M & @10 \\
\hline

BioBART-large & 96.3 & 92.2  & 94.8 & 88.5 & 93.5 &  84.6 \\

BioBART-KP20k & 95.4 & 84.5 & 91.8 & 82.7 & 83.7 & 66.6 \\

BART-KPTimes & 46.0 & 31.2 & 21.4 & 17.4 & 65.8 & 50.7 \\
\hline
\end{tabular}
}
\caption{\label{ext-results}
Extraction percentage in top M and top 10 predictions
}
\end{table}

\section{Conclusion}

This paper introduces \dataset{}, the first large scale dataset for biomedical keyphrase generation. We hope this new dataset will stimulate new research in biomedical keyphrase generation. Several generation models have been trained on this dataset and showed that having more data significantly improves the performances for present and absent keyphrase generation. However, models still perform very poorly on absent keyphrase generation even when using larger amounts of data. In future work, we will focus on how to use \dataset{} to improve biomedical absent keyphrase generation.

\section{Broader Impact and Ethics}

\dataset{} contains some abstracts that are part of copyright protected articles. As the "all rights reserved" statement is optional to be copyright protected, removing articles with this statement does not solve the problem (i.e no copyright statement does not mean free of use data). To be able to collect, work with these data and share the dataset to the research community, we complied with the conditions of US fair use and the exceptions from the 2019/79 EU guideline on using copyright content in text and data mining for research purposes. One of those criteria was to not use the data for commercial purposes which is why we opted for the Creative Commons Non Commercial use license CC-BY-NC v4.0.

\section*{Acknowledgements}
We thank the anonymous reviewers for their valuable input on this article and our colleagues from the TALN team at LS2N for their proofreading and feedback. This work is part of the ANR DELICES project (ANR-19-CE38-0005) and was performed using HPC resources from GENCI-IDRIS (Grant 2022-[AD011013670]).

\bibliography{anthology,custom}

\appendix
\section{Training settings}\label{sec:training_settings}

\begin{itemize}
    \item GPU type: V100 32Go
    \item Number of GPU: 4
    \item Trainer: Seq2SeqTrainer 
    \item Text max size: 512
    \item Reference max size: 128
    \item Optimizer : AdamW
    \item Learning rate: \(5 \times10^{-5}\)
    \item Other hyperparameters: Seq2SeqTrainer default values

\end{itemize}

%
%
%
%

\end{document}